\documentclass{ecai} 
\nolinenumbers

\usepackage{latexsym}
\usepackage{amssymb}
\usepackage{amsmath}
\usepackage{amsthm}
\usepackage{booktabs}
\usepackage{enumitem}
\usepackage{graphicx}
\usepackage{color}
\usepackage{algorithm}
\usepackage{algorithmic}

\newcommand{\BibTeX}{B\kern-.05em{\sc i\kern-.025em b}\kern-.08em\TeX}

\begin{document}


\begin{frontmatter}


\paperid{0545} 


\title{Differential Privacy Image Generation with Reconstruction Loss
and Noise Injection Using an Error Feedback SGD}

\author[A]{\fnms{Qiwei}~\snm{Ma}}
\author[A]{\fnms{Jun}~\snm{Zhang}\thanks{Corresponding Author. Email: zhangjuniris@szu.edu.cn}}

\address[A]{Shenzhen University}


\begin{abstract}
    Traditional data masking techniques such as anonymization cannot achieve the expected privacy protection while ensuring data utility for privacy-preserving machine learning. Synthetic data plays an increasingly important role as it generates a large number of training samples and prevents information leakage in real data. The existing methods suffer from the repeating trade-off processes between privacy and utility. We propose a novel framework for differential privacy generation, which employs an Error Feedback Stochastic Gradient Descent(EFSGD) method and introduces a reconstruction loss and noise injection mechanism into the training process. We generate images with higher quality and usability under the same privacy budget as the related work. Extensive experiments demonstrate the effectiveness and generalization of our proposed framework for both grayscale and RGB images. We achieve state-of-the-art results over almost all metrics on three benchmarks: MNIST, Fashion-MNIST, and CelebA.
\end{abstract}

\end{frontmatter}


\section{Introduction}

Recent advances in machine learning have largely benefited from the massive accessible training data. However, data sharing has raised great privacy concerns, especially in privacy-sensitive scenarios, such as medical records of patients and classroom videos of children.
If the raw data that has not undergone privacy processing is used directly to
train the generative model, it is likely to be attacked in both white-box and black-box settings, 
such as model inversion attacks(MIA) \cite{fredrikson2015Model} and member inference attacks \cite{shokri2017Membership}, 
resulting in the disclosure of private information. The synthesized samples generated 
by Generative Adversarial Networks(GAN) \cite{goodfellow2014Generative} are very similar to its training data. The member 
inference attack \cite{shokri2017Membership} leads to the disclosure of sample privacy information \cite{webster2021This}, 
and the model inversion attack(MIA) can also recover the face sample images used in 
training from the facial recognition system \cite{fredrikson2015Model,khosravyModelInversionAttack2022}. 
In order to deal with the privacy problems in the learning model, traditional 
data masking techniques (i.e., anonymization, 
generalization and perturbation) 
either have poor privacy protection or cannot achieve the expected 
privacy protection effects while ensuring data utility. Nowadays synthetic data plays an increasingly crucial role in enhancing the generalization ability of machine learning models by providing a large number of training samples. Furthermore, synthetic data successfully mitigates the privacy issues inherent in the real data and improves the diversity of the dataset. Generating synthetic data for training in machine learning tasks is anticipated to remain effective in future applications.

The essence of securely synthesizing data is to generate higher-quality data under rigorous privacy constraints such as Differential Privacy \cite{dwork2008Differential}. The synthetic data should be indistinguishable from the original data and can be used for model training. Differential Privacy (DP) is a framework that aims to protect individual data privacy while enabling data analysis. By adding carefully calibrated noise to the training process, DP ensures that the output of the model is not overly dependent on any single data point. One of the most popular techniques to achieve Differential Privacy in deep learning is through the use of Differentially Private Stochastic Gradient Descent (DP-SGD) \cite{abadiDeepLearningDifferential2016}. DP-SGD modifies the traditional gradient descent algorithm by introducing noise\cite{sander2024implicit} into the gradient updates and clipping the gradients to a predefined threshold. This ensures that no single data point can have an excessive influence on the model's training process, thus providing a guarantee of privacy.

However, while DP-SGD is effective, it often introduces challenges such as the need for careful tuning of the noise level and gradient clipping. This is where techniques like Error Feedback come into play. Error feedback\cite{zhang2024differentially} mechanisms are designed to address the bias introduced by gradient clipping. In traditional DP-SGD, gradients are clipped to a fixed norm to prevent them from being too large and potentially leaking information. However, this clipping can distort the gradients, leading to inaccuracies in the model updates. Error feedback accumulates the errors between the clipped and unclipped gradients, compensating for these distortions in subsequent iterations. Therefore, we should keep more details about the original data while maintaining privacy in order to ensure the accuracy of downstream tasks (i.e., classification) based on the generated data. In the field of differential privacy image generation, DP-GAN-DPAC \cite{chen2023private} is the current state-of-the-art method, which showed incredible results on RGB-images such as CelebA dataset \cite{liu2015Deep}, and also improved all metrics on MNIST dataset \cite{lecun1998Gradientbased} and Fashion-MNIST dataset \cite{xiao2017FashionMNIST} used in related works before.

\begin{figure*}[t]
\centering
\includegraphics[width=17.6cm]{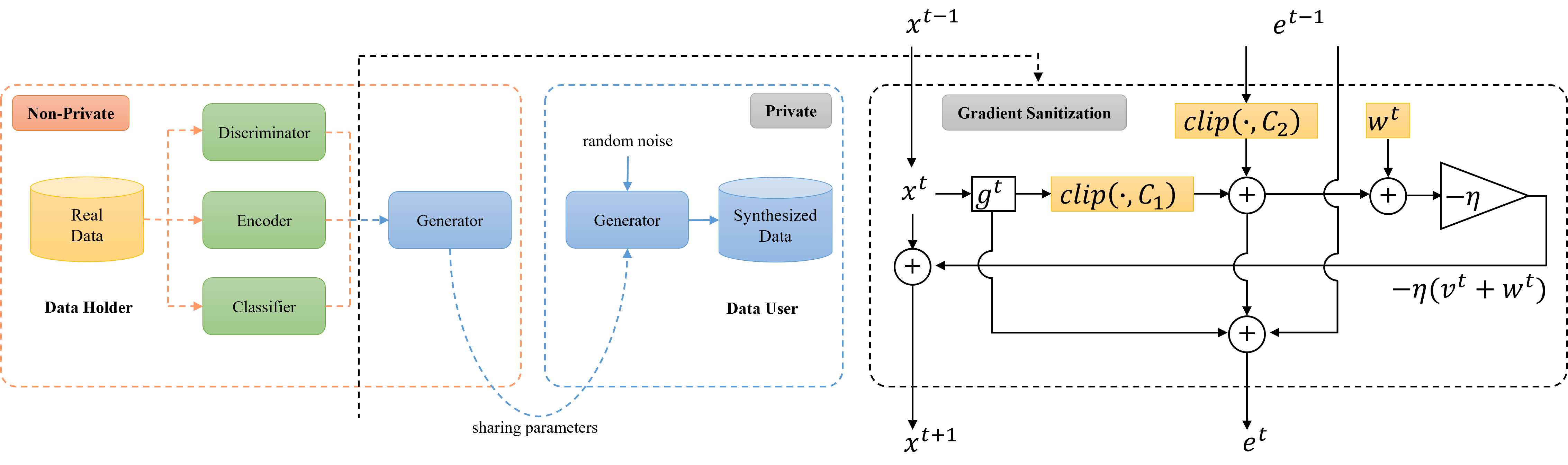}
\caption{Privacy-preserving data synthesis via a differential private generative model. (The left side illustrates the overall framework of our method, while the right side shows the error feedback gradient processing procedure. The algorithm updates the model parameters $x^t$ with $e^{t-1}$, injects the DP noise $w^t$, and computes the clipping error $e^t$ for the next iteration. $C_1$ and $C_2$ are the clipping thresholds for gradient clipping, respectively.)}
\label{fig:model}
\end{figure*}

Based on DP-GAN-DPAC \cite{chen2023private} and StyleGAN \cite{karras2021StyleBased}, we propose a new method of differential privacy image generation. The framework of privacy-preserving data synthesis is illustrated in Figure \ref{fig:model}. In our framework, discriminators and generator will be trained separately, and only the private generator will be published. The gradient sanitization step includes an error feedback mechanism. Reconstruction loss and noise injection are well-established techniques in the generative modeling literature—for example, reconstruction loss in Variational Autoencoders (VAE) \cite{kingma2013AutoEncoding} and noise injection in StyleGAN \cite{karras2021StyleBased}. In our work, we uniquely integrates these two techniques into the differentially private generative training process. Specifically, the reconstruction loss in our framework helps preserve the structural information of original images and provides an additional supervision signal to the generator, which mitigates degradation caused by injected privacy noise. Meanwhile, the noise injected during the upsampling stages enhances image diversity. To the best of our knowledge, we are the first to incorporate these mechanisms into a privacy-preserving image synthesis pipeline, and our experiments demonstrate their effectiveness in improving both data utility and privacy protection compared with previous approaches. 

Our contributions can be summarized as follows:  

\begin{itemize}
    \item We design a framework using a novel stochastic gradient descent algorithm with a clipped error feedback mechanism to eliminate the clipping bias.
    \item We introduce reconstruction loss and  noise injection to improve data utility and diversity.
    \item Experimental evaluations show that our approach outperforms competitors.
\end{itemize}

\section{Related Work}

Privacy-preserving data generation can be broadly categorized into two main approaches: one based on the PATE \cite{papernot2016Semi-supervised} framework and the other leveraging gradient clipping techniques \cite{abadiDeepLearningDifferential2016}.

The PATE framework \cite{papernot2016Semi-supervised} achieves differential privacy by aggregating classification outputs from multiple models (referred to as teachers) using a noisy voting mechanism. Instead of allowing direct access to individual teacher models, a student model learns to predict outputs selected through this differentially private aggregation process. Several works have extended this framework, including PATE-GAN \cite{yoon2018pategan}, G-PATE \cite{long2021gpate}, and DataLens \cite{wangDataLensScalablePrivacy2021}. G-PATE improves upon PATE-GAN by ensuring differential privacy in the information flow between the discriminator and the generator, whereas DataLens introduces a novel noisy gradient compression and aggregation algorithm, TOPAGG, which integrates top-$k$ dimension compression to enhance privacy preservation. These methods provide a structured and controlled approach to privacy-preserving data generation by mitigating direct exposure to individual data instances.

Another approach is gradient clipping techniques, as introduced in DP-SGD \cite{abadiDeepLearningDifferential2016}, provide privacy guarantees by adding random noise to clipped gradients, thereby disrupting the direct transmission of information between the discriminator and generator. Several works have built upon this concept, exploring alternative ways to maintain privacy while improving the utility of generated data. DP-Sinkhorn \cite{cao2021Dona} adopts a transport-based generative method that minimizes the Sinkhorn divergence \cite{Cuturi2013SinkhornDL} in a differentially private manner, eliminating reliance on adversarial networks while maintaining efficient training and deployment. GS-WGAN \cite{chenGSWGANGradientSanitizedApproach2020} introduces a gradient-sanitized Wasserstein GAN that releases only a private generator while non-private discriminators are trained separately to prevent leakage of training data information. By leveraging the Wasserstein-1 metric \cite{arjovskyWassersteinGenerativeAdversarial2017}, GS-WGAN ensures bounded gradients, making the choice of clipping values more manageable and extending applicability to both centralized and decentralized datasets.

The state-of-the-art method, DP-GAN-DPAC \cite{chen2023private}, builds upon GS-WGAN by incorporating a dual-purpose auxiliary classifier trained sequentially using a strategy called sequentialization. Instead of training on mixed real and fake data simultaneously, the classifier alternates between learning from real and generated data separately, thereby improving the quality of generated samples while maintaining strong privacy guarantees. This method also highlights real2gen accuracy (r2g\%) as a key utility metric for future research in privacy-preserving generative models.


\section{Preliminaries}
Differential privacy (DP) \cite{dwork2008Differential} provides a strong privacy guarantee for the algorithms, but a specific and reasonable mean is needed to evaluate the consumption of privacy budget in the training process. That is Rényi Differential Privacy (RDP) \cite{mironov2017Renyi}, and some of the indicators of RDP are also the hyper-parameters that we need to adjust in the DP-SGD \cite{abadiDeepLearningDifferential2016} mechanism. After each gradient descent update, privacy accountant such as moment accountant \cite{abadiDeepLearningDifferential2016} or Renyi Differential Privacy (RDP) accountant \cite{mironov2017Renyi} is used to accumulate privacy costs.

\subsection{Differential Privacy (DP)} 
A randomized function $\mathcal{M}$ gives $\epsilon$ -differential privacy, adding appropriately chosen random noise to the answer of query to avoid leakage of information of participants. If $\mathcal{M}$ with output $S$ of range $\mathcal{R}$, for any adjacent datasets $\mathcal{D}$ and $\mathcal{D}'$, where $\mathcal{D}$ and $\mathcal{D}'$ differ on only one training example, is $(\epsilon, \delta)-DP$ with $\delta$ probability of failing the DP and privacy budgets $\epsilon$, as defined below,

\begin{eqnarray}\label{neighboring_datasets}
    \exists x \in \mathcal{D},\space \mathcal{D} \backslash\{x\}=\mathcal{D}^{\prime}
\end{eqnarray}

\begin{eqnarray}\label{eq:dp}
    Pr[\mathcal{M}(\mathcal{D}) \subseteq \mathcal{S}] \leq e^{\epsilon} Pr\left[\mathcal{M}\left(\mathcal{D}^{\prime}\right) \subseteq \mathcal{S}\right]+\delta
\end{eqnarray}  

If $f(\cdot)$ refers the generator GAN or an arbitrary generating function, the Gaussian sanitization mechanism $\mathcal{M}(\cdot)$ as follows would enable the generator to be differentially private,
\begin{eqnarray}\label{eq:dp_m}
\mathcal{M}(\mathcal{D})=f(\mathcal{D})+\mathcal{N}(0,\left(\sigma \Delta_{2} f\right)^{2})
\end{eqnarray}
where the variance of the Gaussian noise $\mathcal{N}(0,\left(\sigma \Delta_{2} f\right)^{2})$ is determined by the function $f(\cdot)$’s sensitivity value $\Delta_{2} f$. It satisfies
\begin{eqnarray}\label{eq:gaussian}
\Delta_{2} f=\max _{\mathcal{D}, \mathcal{D}^{\prime}}\left\|f(\mathcal{D})-f\left(\mathcal{D}^{\prime}\right)\right\|_{2}
\end{eqnarray}

\subsection{Rényi Differential Privacy (RDP)} 
Compared to DP, RDP supports an easier composition of multiple queries and a clearer privacy guarantee under Gaussian noise. Speciﬁcally, RDP could be easily composed by adding the privacy budget.
In our case, differential cost is consumed step by step. RDP provides convenient composition properties to accumulate privacy cost over a sequence of mechanisms, and that is why we chose RDP accountant for privacy computation as recent works \cite{chen2023private,chenGSWGANGradientSanitizedApproach2020}.

If mechanism $\mathcal{M}$ consists of a sequence of $\mathcal{M}_1, . . . , \mathcal{M}_k$, then for any $i \in [k]$,  $\mathcal{M}_i$ guarantees $(\lambda, {\epsilon}_i)$-RDP, and $\mathcal{M}$ guarantees $(\lambda, \sum_{i=1}^{k} {\epsilon}_i)$-RDP.  $\lambda$ refers to the Rényi order, which is also denoted as $\alpha$ in some literature. And if the Gaussian mechanism is Equation \ref{eq:gaussian}, parameterized by $\sigma$, then $\mathcal{M}$ is $(\lambda, \frac{\lambda \Delta_2 f^2}{2{\sigma}^2})$-RDP.

\subsection{Differential Privacy Stochastic Gradient Descent (DP-SGD) with Error Feedback} 
DP-SGD utilizes a method involving gradient clipping and the addition of random noise to the clipped gradients. However, clipping destroys gradient information and incurs significant utility loss, as reasonable choices of clipping threshold $C$ are significantly lower than the gradient-norms observed. Thus GS-WGAN \cite{chenGSWGANGradientSanitizedApproach2020} use 
Wasserstein-1 metric \cite{arjovskyWassersteinGenerativeAdversarial2017} as the loss function, which is more suitable as the gradients generated are bounded. Here, we need to meet another condition of the loss function, which is that 1-Lipschitz continuous must be satisfied, and then we have a gradient norm close to 1. This allows for the optimal clipping threshold $C$ to be easily obtained. The clipping now is being easy, just replacing the gradient vector $g$ as Equation \ref{eq:gradient} to ensure ${||g||}_2 \le C$.

\begin{eqnarray}\label{eq:gradient}
\hat{{g}}^{t} \leftarrow  \frac{g^{t}}{max(1, \frac{{||g^{t}||}_2}{C})} 
\end{eqnarray}

The final update requires joint noise disturbances to ensure that privacy requirements are guaranteed and the formula can be written as

\begin{eqnarray}\label{eq:update}
x^{t+1}=x^{t}-\eta^{t}(v^{t}+w^{t})
\end{eqnarray}

where $w^t\sim\mathcal{N}(0,\sigma^2\mathbf{I})$. The algorithm first computes the update direction $v^{t}$ by adding the clipped stochastic gradient with the clipped feedback error. Then, the algorithm updates the model parameters $x^{t}$ with $v^{t}$ and injects the DP noise $w^{t}$. And $v^{t}$ is given by
\begin{eqnarray}\label{eq:vt}
v^{t}=clip(g^{(t)}, C_{1})+clip({e}^{t},C_{2}) 
\end{eqnarray}

where the clipping operation  $clip(\cdot)$ has defined by Equation \ref{eq:gradient}.  The additional variable $e^{t}$ tracks the cumulative clipping error at each time step, helping to account for the effects of the clipping operation over the course of the training process. Specifically, $e^{t}$ is updated at each time step using the following recursive formula:
\begin{eqnarray}\label{eq:et}
e^{t}={e}^{t-1}+g^{t-1}-{v}^{t-1}
\end{eqnarray}

This formula for $e^{t}$ describes how the clipping error accumulates over time. At each step $t$, the error $e^{t}$ is updated by adding the difference between the gradient $g^{t-1}$ and the clipped value ${v}^{t-1}$ from the previous step. By tracking and updating this cumulative error, the algorithm ensures that the impact of the clipping operation is properly accounted for, facilitating the stability of the optimization process.

\section{Methodology}
Our method builds on the current state-of-the-art work DP-GAN-DPAC \cite{chen2023private}. Selectively only released generator in GS-WGAN \cite{chenGSWGANGradientSanitizedApproach2020} led us to focus on the optimization of the generator. We introduce Error Feedback SGD, a differential privacy mechanism that involves error accumulation and compensation. Furthermore, we incorporate additional noise injection during the generator's upsampling process and utilize a reconstruction loss with an $L_2$ norm to produce more complex images, carefully balancing the addition of noise with the quality of the images. 

\subsection{Error Feedback SGD} 
This is a novel differential privacy SGD bonded with the Error Feedback mechanism, which accumulates the error between the clipped update to the unclipped one at each iteration and feeds the clipped error back to the next update. So the extra variable $e^{t}$ that records the clipping error is introduced Equation \ref{eq:et}, for recording the deviation caused during the gradient clipping operation and bias compensation when transmitting biased compressed gradients. Certainly, $e^{t}$ have been clipped before being added to the averaged clipped gradient. Using such a clipped Error Feedback mechanism for privacy guarantee, we can balance the functionality of Error Feedback and the DP requirement of the algorithm, and the details about the proof have been shown in \cite{zhang2024differentially}.

Our generative framework is different from the vanilla GAN. There is a multi-party collaboration of generator, discriminators, classifier, and encoder, and the information given by discriminator required for each round of generator updates comes from different discriminators (randomly sampled from pretrained discriminators). Both of these lead to a problem with the accumulation of errors, and a single error is not enough, making it challenging to record the impact of DP operations. We use multiple $e^{t}$ variables to record the errors that occur in the process, and they have a similar calculation process as Equations \ref{eq:error_d}, \ref{eq:error_c} and \ref{eq:error_e}.

\begin{equation}
e^{t}_{d} = {e}^{t-1}_{d} + g^{t-1}_{d} - {v}^{t-1}_{d}
\label{eq:error_d}
\end{equation}

\begin{equation}
e^{t}_{c} = {e}^{t-1}_{c} + g^{t-1}_{c} - {v}^{t-1}_{c}
\label{eq:error_c}
\end{equation}

\begin{equation}
e^{t}_{e} = {e}^{t-1}_{e} + g^{t-1}_{e} - {v}^{t-1}_{e}
\label{eq:error_e}
\end{equation}

These equations describe the cumulative error feedback mechanism that tracks the deviations caused during the gradient clipping operation for different components in the generative framework. Specifically, the errors $e^{t}_{d}$, $e^{t}_{c}$, $e^{t}_{e}$ represent the accumulated clipping errors for the discriminators, classifier, and encoder, respectively. At each time step $t$, the error is updated by adding the difference between the current gradient $g^{t-1}$ and the clipped update $v^{t-1}$ from the previous step. The introduction of these error variables serves to address the challenges posed by the gradient clipping operation. Since clipping can distort the gradients, leading to biases in the updates, it is essential to track the error introduced by this process in order to compensate for it in subsequent iterations. The errors $e^{t}_{d}$, $e^{t}_{c}$ and $e^{t}_{e}$ provide a mechanism for accumulating and correcting these discrepancies for the discriminator, classifier, and encoder, respectively. Each of these components has its own gradient, which is subject to clipping, and hence each requires its own error tracking mechanism. 
  
\subsection{Noise Injection} 

\begin{algorithm}[b]
\caption{Noise Injection}
\label{alg:noise}
\textbf{Input}: Upsampled Image $Y$ \\
\textbf{Parameter}: noise scale $\sigma_{noise}$ \\
\textbf{Output}: Noisy image $Y_{noise}$
\begin{algorithmic}[1]
\FOR{each pixel $(x, y)$ in the image $Y$}
   \STATE Sample noise values $n_i$ from $N(0, \sigma_{noise}^2)$.
\ENDFOR
\STATE Inject noise to the pixel: $Y_{x,y} \leftarrow Y_{x,y} + [n]_{Y_{shape}}$.
\STATE \textbf{return} Noisy image $Y_{noise}$.

\end{algorithmic}
\end{algorithm}

\begin{figure}[b]
\centering
\hspace{8mm}
\includegraphics[width=4cm]{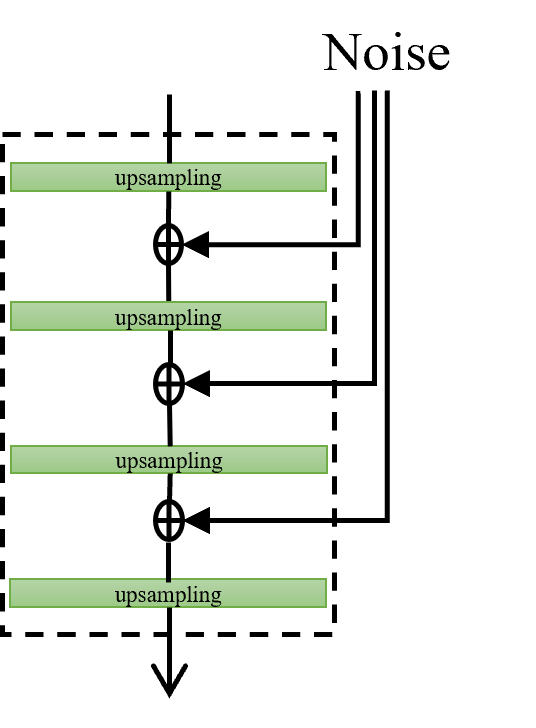}
\caption{the layers of multiple noises adding in}
\label{fig:noise_add_layers}
\end{figure}

In order to improve the performance and enhance the diversity of generated samples, we introduce a strategy of additional noise in the upsampling process of the generator. Specifically, after each upsampling phase of the generator, we inject Gaussian noise with a mean value of $0$ and a standard deviation of $\sigma_{noise}$ into the feature maps by direct addition, as described in Algorithm \ref{alg:noise}. This addition of noise not only increases the randomness of the generated samples but also significantly improves the resolution and quality of the generated images while avoiding schema crashes.
In StyleGAN \cite{Karras2019a,karras2020training}, noise is injected directly into the network to generate images with rich styles, so we add Gaussian noises in upsampling layers after each convolution. It is shown in Figure \ref{fig:noise_add_layers}. We feed a dedicated noise(size of network outputs) to each layer of the synthesis network.

In the network structure of the generator, whenever we perform an upsampling operation, that is, when the size of the feature map is enlarged, we add the generated Gaussian noise to the upsampled feature map element by element before activating the function. This allows the network to upsample while absorbing noisy information, encouraging the generator to maximize the $L_2$ difference between the generated images, to produce images with more diversity and higher visual quality. We will discuss the effect of the choice of noise intensity $\sigma_{noise}$ on the training results in our experiments, and how we balance the relationship between noise injection and image quality.

\subsection{Reconstruction Loss} 
To generate more complex images, we adopt the L2 norm for our reconstruction loss within the Variational Autoencoder (VAE) \cite{kingma2013AutoEncoding}. As the training progresses, we observe that the synthesized images do not retain the same features as the original ones, even though the matrices seem to be good as expected. For example, during training on Fashion-MNIST, the stripes on the clothes are almost completely lost, and we can barely identify the bag in the generated image, as the handle is almost unrecognizable and replaced by a solid black patch. To compute the difference between real and generated samples, we calculate the Euclidean distance and scale it by $\gamma_{recon}$ to prevent model collapse. The Euclidean norm is chosen because it provides a direct measure of the difference between samples, offering a controlled and predictable way to influence the loss function. Unlike Jensen-Shannon divergence, which has a maximum value of 1, the Euclidean norm allows for a more nuanced adjustment of the reconstruction error, making it easier to fine-tune the model without excessive impact due to hyperparameter choices.

The loss function of the discriminator is expressed as follows:

\begin{align}\label{lossD}
\mathcal{L}_{D} & = -\mathbb{E}_{x \sim P}[D(\boldsymbol{x})] 
 + \mathbb{E}_{\tilde{x} \sim Q}[D(\tilde{x})] \nonumber \\
               & \quad + \lambda \mathbb{E}[(\|\nabla D(\alpha x + (1-\alpha) \tilde{x})\|_{2} - 1)^{2}]
\end{align}

where $\mathcal{L}_{D}$ represents the loss function for the discriminator, $D(\boldsymbol{x})$ is the discriminator's output for the real sample $x$, and $\tilde{x}$ is the generated sample. $P$ and $Q$ are the distributions of real and generated samples, respectively. $\lambda$ is the hyperparameter for the gradient penalty term, and $\alpha$ is an interpolation factor. Next, we introduce the classifier loss function:

\begin{align}\label{lossC}
\mathcal{L}_{C} &= - \mathbb{E}_{{z} \sim P_{{z}}}[C(G(z, y), y)]
\end{align}

In this equation, $\mathcal{L}_{C}$ is the classifier's loss, where $C$ represents the classifier, $G(z, y)$ is the image generated by the generator, and $y$ is the class label. $P_{z}$ denotes the prior distribution of the latent code variable $z$.

The encoder's reconstruction loss is expressed as:

\begin{align}\label{lossEn}
\mathcal{L}_{En} &= - \mathbb{E}_{{z} \sim P_{{z}}}[\| G(z) - x \|_2^2]
\end{align}

In this case, $\mathcal{L}_{En}$ represents the encoder's reconstruction loss, where $G(z)$ is the generated image from the latent variable $z$, and $x$ is the original image. The term $|\cdot|_2$ indicates the Euclidean norm.

Finally, the generator's total loss function incorporates the losses from the discriminator, classifier, and encoder:

\begin{align}\label{lossG}
\mathcal{L}_{G} &= -\mathbb{E}_{{z} \sim P_{{z}}}[D(G({z}))] + \lambda_{c} \mathcal{L}_{C} + \gamma_{recon} \mathcal{L}_{En}
\end{align}

In this equation, $\mathcal{L}_{G}$ represents the generator's loss function, which combines the output from the discriminator, the classifier's loss, and the reconstruction loss. The hyperparameters $\lambda{c}$ and $\gamma_{recon}$ are used to weight the classifier and reconstruction losses, respectively.

Note that in the above equation, the same samples and labels, which have been used in the discriminator, are used in the training of the classifier and encoder to avoid additional privacy leakage that is not easily detected. 

\section{Experiments}
\label{sec:Experiments}
\subsection{Experimental Setup}
We conduct experiments on MNIST \cite{lecun1998Gradientbaseda}, Fashion-MNIST \cite{xiao2017FashionMNIST} and CelebA \cite{liu2015Deep} dataset over quality and utility evaluation metrics. Specifically, MNIST and Fashion-MNIST datasets both contain 60000 training examples and 10000 validation examples of 28 × 28 grey-scale images consisting of 10 labels. The CelebA dataset contains 202599 color images of celebrity faces, each with 40 attribute annotations. As the current state-of-the-art work DP-GAN-DPAC did, we only take the binary “gender” attribute as the label and resize the images to 32 × 32 × 3. The data is partitioned into three subsets: 162770 training, 19867 validation, and 19962 test. During all training and inference processes, labels from origin datasets are only used in unreleased components, like discriminator. Labels generated from a uniform prior distribution, which is independent of the training dataset, thus do not incur additional privacy costs.
Before training, we partition the dataset into $k$ subsets, $D_1, ..., D_k$.

\subsection{Evaluation Metrics} 
To evaluate output quality and utility, we use two metrics: sample quality and usefulness for downstream tasks,  which are also used in this line of work. The sample quality includes Inception Score(IS) \cite{heusel2017GANs,salimans2016Improveda} and Frechet Inception Distance(FID) \cite{li2017ALICE}, both of which are benchmark metrics in GAN researches. The \textbf{Inception Score} evaluates the quality and diversity by examining the classifier's confidence in assigning class labels to the generated images and the evenness of the distribution across classes. A higher IS implies better quality and diversity. The \textbf{Frechet Inception Distance}, on the other hand, measures the discrepancy between the feature distributions of real and generated images by calculating the statistical distance between the estimated Gaussian distributions of their features. A lower FID score is associated with higher quality, as it suggests a smaller difference between the real and generated image distributions. 

To evaluate utility, the standard is to use the g2r\%(gen2real accuracy), which computes the accuracy on real data of a downstream classifier trained using generated data, where for the classifier, Multi-Layer Perceptron (MLP) and Convolutional Neural Network (CNN) are both used. 
In our experiments, we follow GS-WGAN \cite{chenGSWGANGradientSanitizedApproach2020} and DP-DPAC-GAN \cite{chen2023private} and use the implementation for experimental evaluation.

\begin{algorithm}[t]
\caption{Differential Privacy Image Generation via Multi-Component Training}
\label{alg:algorithm}
\textbf{Input}: Dataset $D$ \\
\textbf{Parameter}: Subsampling rate $\gamma$, noise scale $\sigma$, the number of discriminator $n_{dis}$, training iterations $T$, learning rates ${\eta}_D$, ${\eta}_C$ and ${\eta}_G$, batch size $B$, class label $y$\\
\textbf{Output}: Private generator  $\boldsymbol{\theta}_G$
\begin{algorithmic}[1] 
\STATE Load pre-trained discriminators $\boldsymbol{\theta}_{D}^{k}$, initialise private generator  $\boldsymbol{\theta}_G$.
\FOR{$k\leftarrow 1$}
\STATE sample subset index k with batch size $B$, initialise classifier ${\theta}_C$

\FOR{$n_{dis}\leftarrow 1$}
\STATE Sample batch $\{x_i\}_{i=1}^B \subseteq {D_k}$\;
\STATE Sample batch $\{z_i\}_{i=1}^B \ with \ {z_i\sim P_z}$\;
\STATE Compute mean discriminator gradient $g_D$\;  
\STATE $\boldsymbol{\theta}_D^k \leftarrow  \boldsymbol{\theta}_D^k - {\eta}_D \cdot  \boldsymbol{g}_D $

\ENDFOR
\FOR{$n_{en}\leftarrow 1$}
\STATE Sample batch $\{x_i\}_{i=1}^B \subseteq {D_k}$\;
\STATE Sample batch $\{z_i\}_{i=1}^B \ with \ {z_i\sim P_z}$\;
\STATE Compute mean encoder gradient $g_E$\;
\STATE $\boldsymbol{\theta}_E^k \leftarrow  \boldsymbol{\theta}_E^k - {\eta}_E \cdot  \boldsymbol{g}_E $

\ENDFOR
\FOR{$n_{f}\leftarrow 1$}
\STATE Sample batch $\{z_i\}_{i=1}^B \ with \ {z_i\sim P_z}$\;
\STATE Compute mean classifier gradient $g_C$\;
\STATE $\boldsymbol{\theta}_C^k \leftarrow  \boldsymbol{\theta}_C^k - {\eta}_C \cdot  \boldsymbol{g}_C $
\ENDFOR

\FOR{$n_{r}\leftarrow 1$}
\STATE Sample batch $ \{x_i\}_{i=1}^B \subseteq {D_k}$\;
\STATE Compute mean classifier gradient $g_{C^{'}}$\;
\STATE $\boldsymbol{\theta}_C^{k} \leftarrow  \boldsymbol{\theta}_C^{k} - {\eta}_C \cdot  \boldsymbol{g}_{C^{'}}$\;
\ENDFOR

\STATE Compute mean sanitized generator gradient $g_G$\;
\STATE $\boldsymbol{\theta}_{G} \leftarrow \boldsymbol{\theta}_{G}-\eta_{G} \cdot \tilde{\boldsymbol{g}}_{G}$\;
\STATE Accumulate privacy cost $\epsilon$\;
\ENDFOR

\STATE \textbf{return} Generator $\boldsymbol{\theta_G}$, privacy cost $\epsilon$
\end{algorithmic}
\end{algorithm}

\subsection{Implementation Details.} 
Same as previous works \cite{chen2023private, chenGSWGANGradientSanitizedApproach2020}, we have used DCGAN for the discriminator and classifier, and ResNet (adapted from BigGAN) for the generator. Our encoder comes from the encoder of vanilla VAE \cite{kingma2013AutoEncoding}. 
The clipping operation  $clip(\cdot)$ has been defined by Equation \ref{eq:gradient}. $e^{t}$ is an extra variable that records the clipping error, which satisfies Equation \ref{eq:et}.
Other implementation details are same as the related works \cite{chen2023private, chenGSWGANGradientSanitizedApproach2020}. We have also employed the subsampling technique, which strategically reduces the size of the dataset by selecting a representative subset of data points, thereby minimizing the exposure of sensitive information. Additionally, to improve the performance and convergence speed of our model, we have pre-trained the discriminators before commencing the main training phase. This pre-training step allows the discriminators to develop an initial understanding of the data distributions, which in turn facilitates a more effective and efficient learning process during the subsequent training iterations. We implement the gradient sanitization mechanism by registering a backward hook to the selected portion of generator gradient $ \bigtriangledown_{G(z)}\mathcal{L}_{G}(\theta_G) $. We adopt the official implementation of \cite{wang2019Subsampled} for accumulating the privacy costs after each generator iteration. Due to the introduction of error feedback, it is necessary to adjust the intensity of noise addition, which we refer to the specific implementation of \cite{zhang2024differentially}. It satisfies $w^t\sim\mathcal{N}(0,\sigma^2 C_{1}^2 (1 + 2C_2))$, where $\sigma$ is the noise multiplier, affected by the number of training iterations and $C_1$, $C_2$ are the clipping thresholds of gradient and error variant $e^t$. 

The value choice of differential privacy $\delta = 1e-5$ and $\epsilon=10$ were both kept the same as in \cite{chen2023private, chenGSWGANGradientSanitizedApproach2020}, to facilitate a fair comparison. Through a large number of experiments, we have selected the noise injection coefficient $\sigma_{noise}=0.1$ and the reconstruction coefficient $\gamma_{recon}=1.0$ using grid search. The details of our algorithm is shown in Algorithm \ref{alg:algorithm}. 

\subsection{Comparison with Baselines}

\begin{figure}[h]
\centering
\begin{tabular}{lll}
\hline \hline
Method      & \multicolumn{1}{c}{MNIST}\\ \hline
GS-WGAN     &   \raisebox{-.5\height}{\includegraphics[width=5.5cm]{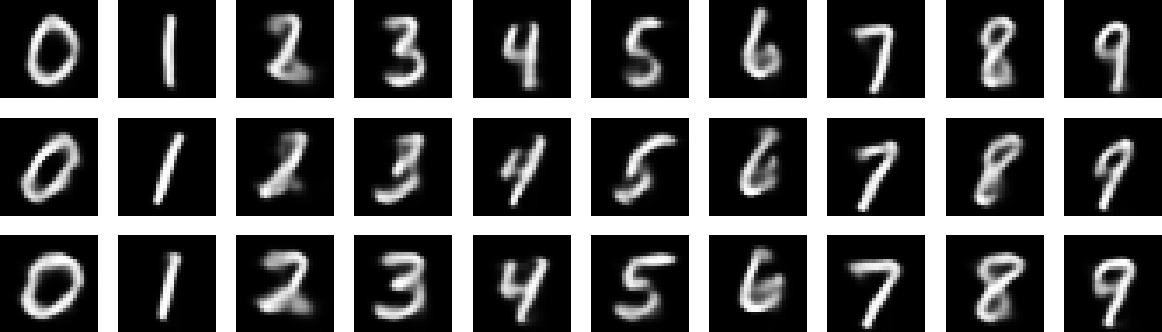}} \\ \\
DP-GAN-DPAC &  \raisebox{-.5\height}{\includegraphics[width=5.5cm]{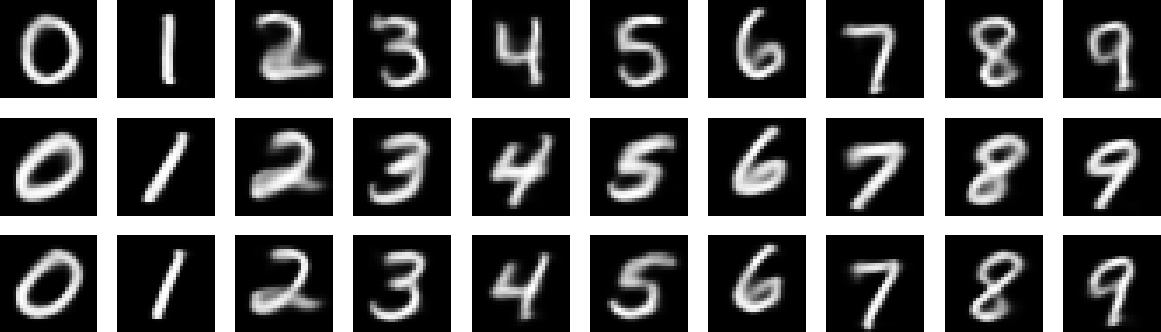}} \\ \\ 
Ours        &  \raisebox{-.5\height}{\includegraphics[width=5.5cm]{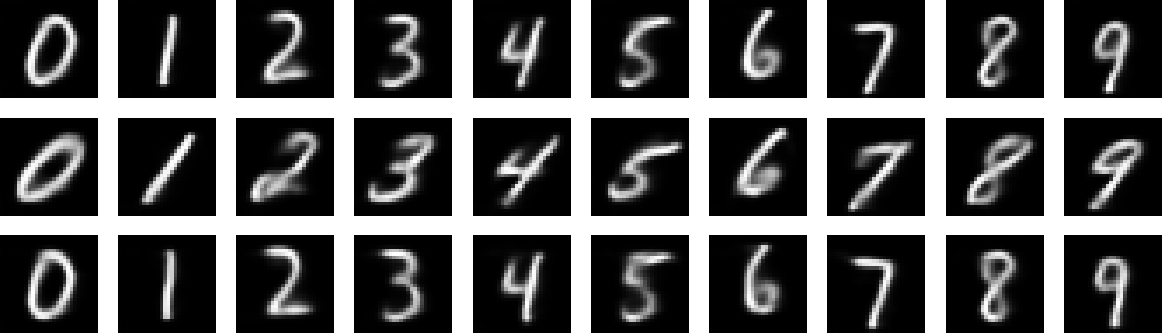}} \\
\hline \hline
\end{tabular}
\caption{Images generated for MNIST.}\label{results_mnist}
\end{figure}

\begin{figure}[h]
\centering
\begin{tabular}{lll}
\hline \hline
Method      & \multicolumn{1}{c}{Fashion-MNIST}  \\ \hline
GS-WGAN     &  \raisebox{-.5\height}{\includegraphics[width=5.5cm]{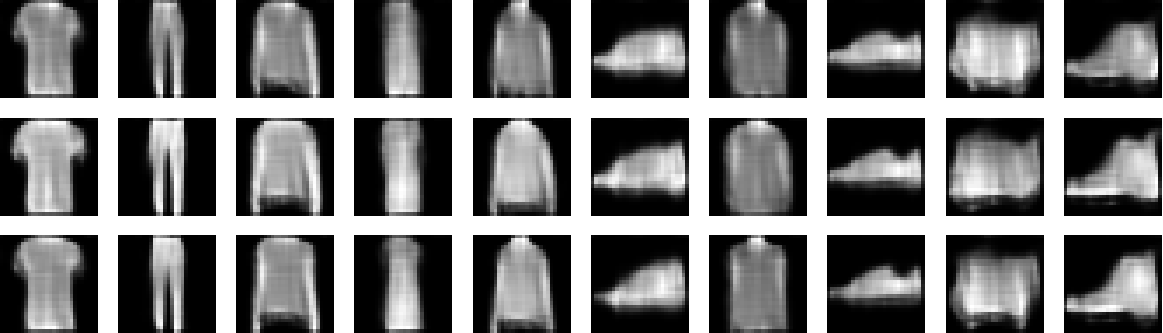}} \\ \\
DP-GAN-DPAC &  \raisebox{-.5\height}{\includegraphics[width=5.5cm]{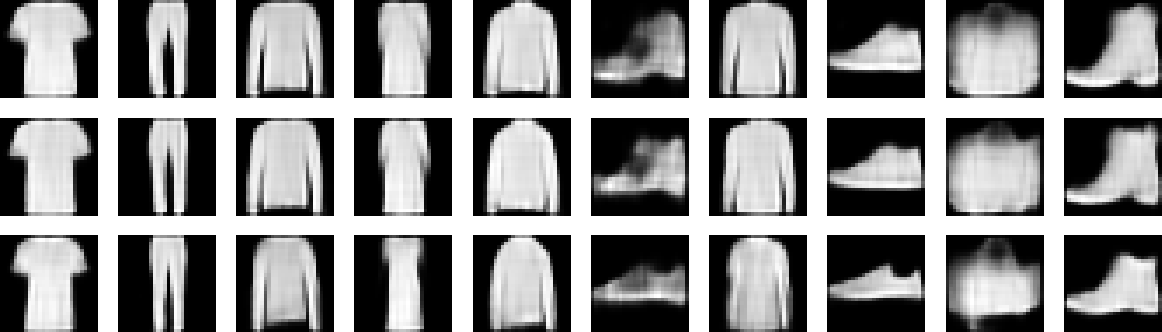}} \\ \\
Ours        &  \raisebox{-.5\height}{\includegraphics[width=5.5cm]{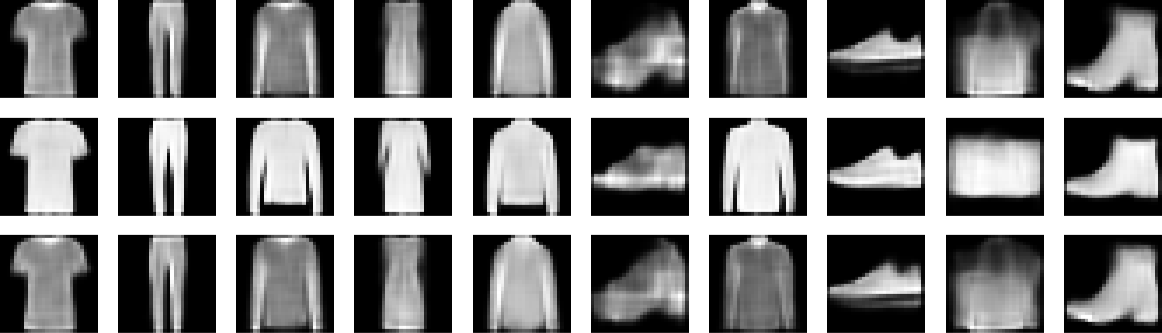}} \\
\hline \hline
\end{tabular}
\caption{Images generated for Fashion-MNIST. In the penultimate column, with clear handles, the bags in our results are close to the real, and in the rest columns, the blank space at the neck of the T-shirts and the sleeves of the skirts are visible.}
\label{results_fashionmnist}
\end{figure}

As shown in Figure \ref{results_mnist}, Our approach generated the most visually pleasing outcomes across all evaluated datasets compared to several state-of-the-art benchmarks. For MNIST, the improvements in results are not obvious. The reason for this is that MNIST is relatively straightforward to generate, and recent works have already achieved commendable results. However, we can see significant improvements in our approach to the more challenging Fashion-MNIST and CelebA dataset. Examples of synthetic images for Fashion-MNIST from various methods are presented in Figure \ref{results_fashionmnist}. Below are examples of synthetic images for CelebA (Figure \ref{results_CelebA})generated by various methods.

On the more challenging CelebA dataset, GS-WGAN \cite{chenGSWGANGradientSanitizedApproach2020} can resemble faces, but with many mosaics on the faces, it is hard to visualize gender differences. DP-GAN-DPAC \cite{chen2023private} can show clear signs of gender. However, the outputs are much blurrier compared to ours. Our images contain more details and create rich visuals, which the latter can be attributed to the noise injection mechanism we proposed. In addition, we intuitively believe that our method does not reduce the blurriness of the image too much, partly because the images used in the experiment are only 32x32. On the other hand, the injection of noise may limit the ability of the model to generate clearer images. Even so, it is still obvious that our generated image effect is better than the former.

\begin{table}[]
\caption{Comparing IS ↑ and FID ↓ on various datasets.}
\centering
\setlength{\tabcolsep}{1mm}     
\begin{tabular}{l|ll|ll|ll}

\hline
\multicolumn{1}{l|}{} & \multicolumn{2}{c|}{MNIST}& \multicolumn{2}{c|}{F-MNIST} & \multicolumn{2}{c}{CelebA} \\
\multicolumn{1}{l|}{Method}  & \multicolumn{1}{c}{IS}   & \multicolumn{1}{c|}{FID}   & \multicolumn{1}{c}{IS}  & \multicolumn{1}{c|}{FID}    & \multicolumn{1}{c}{IS}  & \multicolumn{1}{c}{FID}  \\ \hline

Real  & 9.80    & 1.02  &8.98  &1.49& 2.77& 1.06\\
GS-WGAN & \multicolumn{1}{c}{9.23} & \multicolumn{1}{c|}{61.34} & 5.32& 131.34& 1.85& 187.36\\
DPSinkhorn  & \multicolumn{1}{c}{-}& 55.56& \multicolumn{1}{c}{-}& 129.40& \multicolumn{1}{c}{-}& 168.40\\
DP-GAN-DPAC  & 9.71& 54.06& 6.60& 90.77& 1.90& 139.99\\
Ours& \textbf{9.74}& \textbf{49.41}&\textbf{6.72}  & \textbf{83.48} & \textbf{2.26} & \textbf{114.03} \\ \hline 
\end{tabular}

\label{results}
\end{table}

Our method still gets the best performances for IS and FID metrics, as Table \ref{results}. As MNIST and Fashion-MNIST are relatively easy to generate, the advantages of our model are not distinct. To compare with the baseline works, we do a lot of validation on MNIST and Fashion-MNIST. However, we strongly recommend that future work focus on complex and diverse image generation or rigorous privacy guarantee. Due to the extensive efforts of previous work, such endeavors have been able to achieve quite satisfactory results; on the other hand, based on the evaluation conducted on real datasets as shown in Table \ref{results}, the quality of the generated samples appears to have reached the bottleneck. 

\begin{table}[t]
\caption{Comparing gen2real accuracy ↑ on various datasets.}
\centering
\setlength{\tabcolsep}{1mm} 
\begin{tabular}{l|cc|cc|cc}
\hline  & \multicolumn{2}{|c|}{ MNIST } & \multicolumn{2}{c|}{ F-MNIST } & \multicolumn{2}{c}{ CelebA } \\
Method  $\uparrow$  & MLP & CNN & MLP & CNN & MLP & CNN \\
\hline  
GS-WGAN & 0.79& 0.80& 0.65& 0.65& 0.68& 0.66 \\
DPSinkhorn& 0.80& 0.83 & 0.73 & 0.71  & 0.76 & 0.76 \\
DP-GAN-DPAC \footnotemark[1]
& 0.82& 0.84& \textbf{0.74}&\textbf{0.71}& \textbf{0.80}& 0.83\\
Ours &  \textbf{0.86}  & \textbf{0.85}  & \textbf{0.74}  & 0.70 &  \textbf{0.80}  &   \textbf{0.88} \\
\hline 
\end{tabular}

\label{table_g2r}
\end{table}

\begin{figure}[tb]
\centering
\setlength{\tabcolsep}{1pt}
\begin{tabular}{c|c|c}
GS-WGAN & DP-GAN-DPAC & Ours \\ \hline
    \raisebox{-.1\height}{\includegraphics[height=10.5cm]{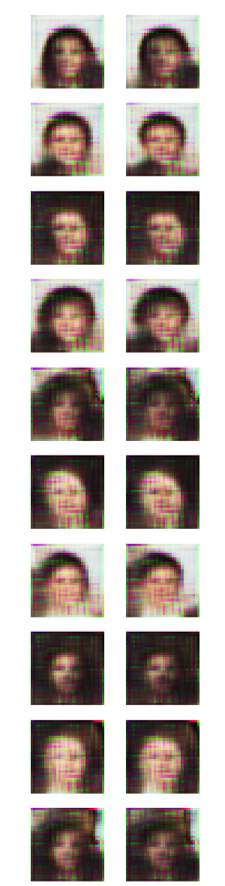}}    &        \raisebox{-.1\height}{\includegraphics[height=10.5cm]{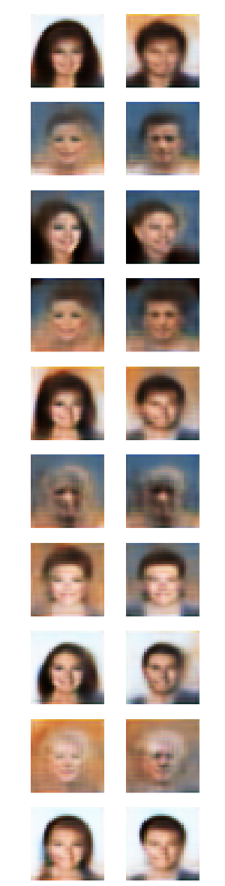}}     &    \raisebox{-.1\height}{\includegraphics[height=10.5cm]{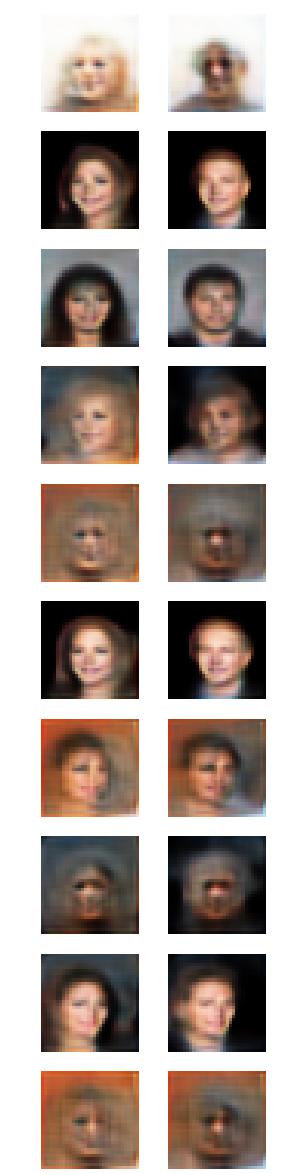}} 
\end{tabular}
\caption{Images generated for CelebA. There are several individuals in our results (all of these are experimental results and have not been manually selected), whereas, in the DP-GAN-DPAC, many results are even completely identical. At the same pixel level, our results excel in detail, especially in the second and third rows. Zoom in for the best view.}
\label{results_CelebA}
\end{figure}

\subsection{Ablation Experiments}
In Table \ref{table_noise} and \ref{table_recon}, we use different noise $\sigma_{noise}$ and reconstruction loss $\gamma_{recon} $ and compare the baseline work, DP-GAN-DPAC \cite{chen2023private}, the first line(base). All parameters are as consistent as possible with baseline work to eliminate additional effects. We can see that the noise injection and reconstruction loss used in our framework have significantly improved the experimental results. We demonstrate the performance of noise injection and reconstruction loss through extensive experiments. As illustrated in Table \ref{table_noise}, Noise injection improves the results to varying degrees, but too much noise may render the results ineffective. Moreover, the effect of the introduction of refactoring loss is particularly significant. By employing the encoder, both IS and FID metrics surpass the baseline results on nearly all datasets, even when all other settings remain identical.

\begin{table}[h]
\caption{Comparing different $\sigma_{noise}$}
\centering
\setlength{\tabcolsep}{1mm} 
\begin{tabular}{l|cc|cc|cc}
\hline  & \multicolumn{2}{|c|}{ MNIST } & \multicolumn{2}{c|}{ F-MNIST } & \multicolumn{2}{c}{ CelebA } \\
Noise  & IS & FID & IS & FID & IS & FID \\
\hline  
0.0(base) & \textbf{9.71} & 54.6 & 6.60 & 90.77 & 1.90 & 139.99 \\
0.1 & 9.60 & 49.41 & \textbf{6.77} & \textbf{85.58} & 2.37 & \textbf{97.70} \\
1.0 & 9.27 & \textbf{46.84} & 6.36 & 102.62 & \textbf{2.98} & 218.38 \\
\hline 
\end{tabular}

\label{table_noise}
\end{table}

\begin{table}[h]
\caption{Comparing different $\gamma_{recon} $}
\centering
\setlength{\tabcolsep}{1mm} 
\begin{tabular}{l|cc|cc|cc}
\hline  & \multicolumn{2}{|c|}{ MNIST } & \multicolumn{2}{c|}{ F-MNIST } & \multicolumn{2}{c}{ CelebA } \\
Reconstruction&IS&FID&IS&FID&IS&FID\\
\hline    
0.0(base) & 9.71 & 54.6 & 6.60 &  \textbf{90.77} & 1.90 & 139.99 \\
1.0 & \textbf{9.74} &  \textbf{51.63} &  \textbf{6.63} & 94.99 &  \textbf{2.06} & \textbf{99.10} \\
\hline 
\end{tabular}

\label{table_recon}
\end{table}

\footnotetext[1]{The results of DP-GAN-DPAC are recorded after running the source code provided in the original paper.}

We have tested which iteration to introduce reconstruction loss and noise injection. No matter which iteration they are introduced, there is an immediate improvement in the effect of IS and FID, which is a direct proof to the effectiveness of the components we added. 

\section{Conclusion}

In this paper, a novel error feedback-based stochastic gradient descent method has been utilized in our privacy generation framework, which has significantly enhanced the quality of the generation. We consider the introduction of reconstruction loss to have a similar distribution with the origin as possible, and the reconstruction loss is used to assist in data generation at a certain stage of training. In the prior work, we could intuitively observe that some details were missing in the generated data as opposed to the original data, which was highly detrimental to further data utilization. Consequently, we mitigated the loss of information in the images by introducing noise injection, thereby offering more diversity. We have also experimentally demonstrated the importance of the above approaches, and we consider that appropriate prior noise can be added for specific tasks to generate more usable data in future privacy generation tasks.



\begin{ack}
By using the \texttt{ack} environment to insert your (optional) 
acknowledgements, you can ensure that the text is suppressed whenever 
you use the \texttt{doubleblind} option. In the final version, 
acknowledgements may be included on the extra page intended for references.
\end{ack}



\bibliography{mybibfile}

\end{document}